\documentclass[runningheads]{llncs}
\usepackage[T1]{fontenc}
\usepackage{graphicx,verbatim}
\usepackage{booktabs}
\usepackage{multirow}
\usepackage{subcaption}
\usepackage{amsmath}
\usepackage{amssymb}
\usepackage{amsfonts}

\begin{document}
\title{CURE: A Multimodal Benchmark for Clinical Understanding and Retrieval Evaluation}
%

\author{
Yannian Gu\inst{1}\thanks{These authors contributed equally.} \and
Zhongzhen Huang\inst{1}\protect\footnotemark[1] \and
Linjie Mu\inst{1} \and
Xizhuo Zhang\inst{1} \and
Shaoting Zhang\inst{2}$^{\dagger}$ \and
Xiaofan Zhang\inst{1,3}$^{\dagger}$
}

\authorrunning{Y. Gu et al.}

\institute{
Shanghai Jiao Tong University, Shanghai, China \and
SenseTime Research, China \and
Shanghai Innovation Institute, Shanghai, China \\
\email{
xiaofan.zhang@sjtu.edu.cn
}
}

\maketitle              

\begin{abstract}
Multimodal large language models (MLLMs) demonstrate considerable potential in clinical diagnostics, a domain that inherently requires synthesizing complex visual and textual data alongside consulting authoritative medical literature. 
However, existing benchmarks primarily evaluate MLLMs in end-to-end answering scenarios. 
This limits the ability to disentangle a model’s foundational multimodal reasoning from its proficiency in evidence retrieval and application. 
We introduce the Clinical Understanding and Retrieval Evaluation (CURE) benchmark. 
Comprising $500$ multimodal clinical cases mapped to physician-cited reference literature, CURE evaluates reasoning and retrieval under controlled evidence settings to disentangle their respective contributions. 
We evaluate state-of-the-art MLLMs across distinct evidence-gathering paradigms in both closed-ended and open-ended diagnosis tasks. 
Evaluations reveal a stark dichotomy: while advanced models demonstrate clinical reasoning proficiency when supplied with physician reference evidence (achieving up to $73.4\%$ accuracy on differential diagnosis), their performance substantially declines (as low as $25.4\%$) when reliant on independent retrieval mechanisms. 
This disparity highlights the dual challenges of effectively integrating multimodal clinical evidence and retrieving precise supporting literature.
CURE is publicly available at \url{https://github.com/yanniangu/CURE}.

\keywords{Multimodal Case Retrieval \and Clinical Reasoning Benchmark}
\end{abstract}

\section{Introduction}
Multimodal large language models (MLLMs) have recently shown impressive capabilities in jointly modeling vision and language, enabling progress in tasks that require grounded interpretation across images and narratives~\cite{li2023llava,sellergren2025medgemma,huang2025elicit,qin2026incentivizing}.
In medicine, this progress is especially compelling: clinical diagnosis rarely depends on a single modality.
Clinicians synthesize heterogeneous cues (e.g., imaging patterns, temporality, comorbidities, laboratory trends) to form differential diagnoses and decide next steps~\cite{huang2020fusion,zhang2025multimodal}.
Critically, when cases are ambiguous or high-stakes, clinicians often consult authoritative diagnostic literature, such as textbooks, guidelines, or peer-reviewed case reports, to justify interpretations, refine differentials, and support downstream decisions such as treatment planning and prognosis estimation~\cite{del2014clinical}.
As a result, \emph{tracing the supporting evidence} is not merely an auxiliary step, but a core component of reliable clinical reasoning.

\begin{figure}[t]
    \centering
    \includegraphics[width=\textwidth]{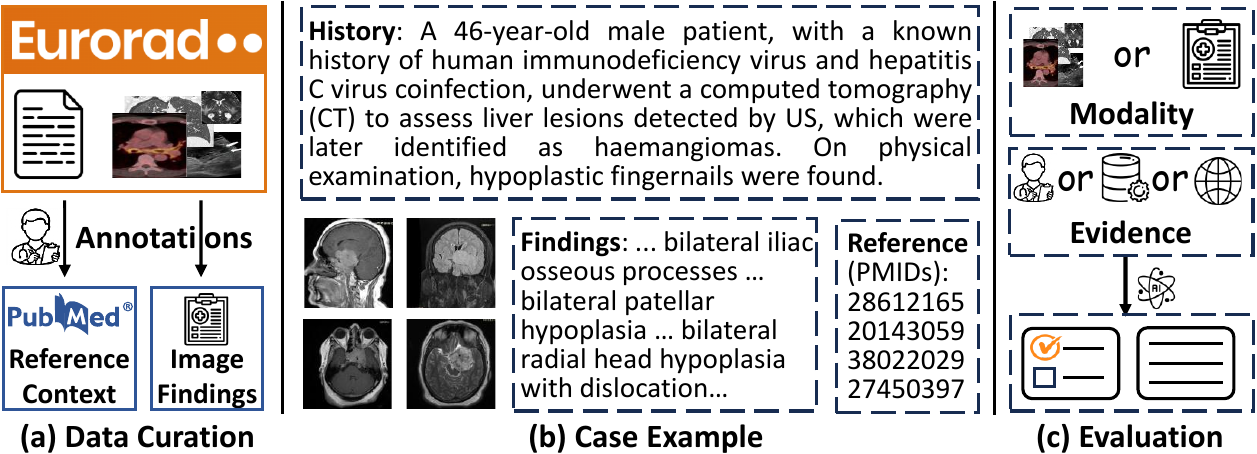}
    \caption{CURE benchmark pipeline. Cases are paired with PubMed reference evidence, and MLLMs are evaluated under different retrieval paradigms to disentangle multimodal understanding from evidence-grounded diagnosis performance.}
    \label{fig:cure_pipeline}
\end{figure}

This perspective has motivated a growing body of work that augments MLLMs with external knowledge through retrieval-augmented generation (RAG)~\cite{lewis2020retrieval,xiong2024benchmarking} and, more recently, agent-based frameworks~\cite{team2025mirothinker,xu2025comprehensive} that iteratively search, read, and refine.
Despite rapid methodological iteration, evaluating these systems presents a dual challenge.
On the \emph{system level}, existing agent pipelines rely on heterogeneous search corpora (e.g., web sources, mixed medical repositories, proprietary databases) and entangled implementation choices, making results hard to interpret and reproduce across settings.
On the \emph{benchmark level}, established medical datasets such as MedXpertQA~\cite{zuo2025medxpertqa} and MMMU~\cite{yue2024mmmu} have advanced MLLM evaluation across visual question answering and multimodal reasoning, but they primarily assess end-to-end answering in isolation, scoring models solely on their final responses to given inputs. 
They lack a systematic testbed for evidence-grounded retrieval against a unified, authoritative reference set.
This leaves a key gap: developers lack an objective and clinically faithful way to \emph{disentangle} whether an incorrect answer arises from deficient multimodal understanding, retrieval failure, or inability to leverage retrieved evidence. 

To address this gap, we introduce the \textbf{Clinical Understanding and Retrieval Evaluation (CURE)}, a benchmark designed to evaluate MLLMs under a clinically motivated setting where models are expected to (i) interpret multimodal clinical inputs, (ii) retrieve relevant evidence from a unified, authoritative diagnostic literature source, and (iii) generate diagnoses that are both accurate and evidence-grounded.  CURE is built to explicitly decouple model capability from retrieval behavior by providing standardized reference content and evaluation protocols that isolate where systems succeed or fail.

Using CURE, we benchmark state-of-the-art MLLMs which reveals a striking contrast: while advanced models demonstrate strong clinical reasoning when supplied with physician reference evidence (achieving up to 73.4\% accuracy on open-ended diagnosis), their end-to-end performance drops sharply (dropping to 25.4\%) when tasked with independent retrieval. 
This massive performance gap highlights that the critical bottleneck in current clinical AI is no longer isolated medical reasoning, but rather the ability to effectively translate multimodal visual cues into precise literature retrieval.
In summary, our contributions are threefold:
\begin{itemize}
\item We build a multimodal evidence-retrieval benchmark, a 500-case dataset with imaging and corresponding findings, grounded by physician-cited references as standard external retrieval context.
\item We evaluate SOTA MLLMs across retrieval paradigms (dense and agent-based) and analyze multimodal understanding of medical images in both closed-ended and open-ended setting diagnosis.
\item We provide a comprehensive analysis of failure modes, offering quantitative evidence of persistent retrieval bottlenecks and highlighting where current systems fail to retrieve clinically decisive literature.
\end{itemize}

\section{The CURE Benchmark}

We construct the \textbf{CURE} benchmark, which transitions the evaluation paradigm from simple visual question-answering to evidence-based diagnostic reasoning, bridging the gap between semantic matching and authoritative clinical inference.

\subsection{Data Collection and Benchmark Setup}

To construct CURE, we systematically collected all 530 clinical cases published in Eurorad during 2025. To simulate a real-world diagnostic starting point, we isolated the \textit{Clinical History} as the query text and strictly removed all diagnostic answers, retrospective analyses, and metadata. Images were retained in their native formats and resolutions.
We also extracted the Imaging Findings authored by the radiologist as an optional textual reference for image understanding, by removing any explicit diagnostic labels or answer-revealing phrases and excluded overlapping retrospective sections, yielding observation-level descriptions rather than diagnostic conclusions.
Unlike benchmarks relying on LLM-generated summaries, CURE establishes a rigorous Ground Truth (GT) using explicitly physician-cited literature. We extracted the PubMed IDs (PMIDs) natively provided in each Eurorad report and mapped them deterministically against a custom offline database of 39.65 million PubMed abstracts. This ensures models are evaluated solely on their ability to surface the exact scientific evidence utilized by human experts.
Finally, we partitioned the dataset into a development set (30 cases, stratified by specialty and modality) for pipeline tuning, and a strictly isolated test set (500 cases) for zero-shot evaluation. Because all cases were published in 2025, they post-date the knowledge cutoffs of current MLLMs, effectively mitigating data contamination.

\subsection{Dataset Statistics}

Statistical analysis of the CURE benchmark reveals distinct characteristics:

\begin{itemize}
    \item \textbf{Unbiased Demographic Representation:} The dataset exhibits near-perfect gender parity (50.5\% male, 49.5\% female). Furthermore, it spans a comprehensive age distribution ranging from neonates (0 years) to the elderly (94 years), with a mean age of 41.9 years. This demographic balance helps mitigate algorithmic bias regarding age and gender.

    \item \textbf{High Multimodal Complexity:} Cases span 11 major medical specialties, predominantly featuring Musculoskeletal (19.5\%), Neuroradiology (18.5\%), and Abdominal imaging (16.1\%). The imaging modalities are highly diverse, dominated by CT (31.1\%) and MR (29.9\%), followed by Ultrasound (12.6\%). The information density per case is exceptionally high: an average case presents the MLLM with \textbf{8.0 medical images} and requires the retrieval of \textbf{6.5 distinct PubMed abstracts} as evidence.
    
    \item \textbf{Differential Diagnosis (DDx) Complexity:} In the close-ended setting, models are challenged with a highly targeted list of plausible diagnoses. The number of candidate options per case ranges from 4 to 5, with a mean of 4.7 options. This establishes a baseline random-guess accuracy of approximately 21.3\%, demanding strong deductive reasoning from the models.
\end{itemize}

\subsection{Task Formulation: End-to-End Diagnostic Reasoning}

Each case is defined as a multimodal query $Q=(T,V)$, comprising clinical text $T$ and a set of images $V=\{v_i\}_{i=1}^n$. We evaluate end-to-end diagnostic reasoning under two distinct settings:

\paragraph{1. Differential Diagnosis Selection:} Models receive $Q$ alongside 4--5 candidate diagnoses derived from the original Eurorad report. GPT-5.2 was utilized to standardize terminologies and generate clinically plausible distractors when necessary. A board-certified physician reviewed a random sample of 50 cases to guarantee distractor validity. Accuracy (ACC) is used as the primary metric.

\paragraph{2. Open-ended Diagnosis Prediction:} Models receive solely $Q$ and must autonomously generate a ranked list of diagnoses. Performance is measured via Hit@1 and Hit@3, which represent whether the correct answer appears in the top-1 or top-3 ranked results, respectively. To account for semantic variations in clinical expressions, we employ DeepSeek-V3.1 as an LLM-as-a-judge to evaluate alignment with the ground truth. This automated evaluation demonstrates high reliability, achieving strong agreement (Cohen's $\kappa = 0.86$) with a blinded physician review of 50 sampled cases.

\subsection{Data Availability and Reproducibility}

To ensure complete reproducibility, the CURE benchmark, including prompts, predefined splits, evaluation scripts, and offline PubMed retrieval codebase, will be publicly released on HuggingFace under the CC BY-NC-SA 4.0 license upon publication.
All data collection procedures comply with Eurorad's terms of use, and since the benchmark utilizes pre-anonymized, publicly available educational reports, it meets the criteria for ethical exemption from institutional review.

\section{Experiments}
\label{sec:experiments}

\subsection{Experimental Setup}
\label{subsec:exp_setup}

We benchmark a diverse suite of proprietary models (GPT-5.2~\cite{singh2025openai}, Gemini-3-Pro-Preview) and open-weight models (Kimi-K2.5~\cite{team2026kimi}, GLM-4.6V~\cite{vteam2025glm45vglm41vthinkingversatilemultimodal}, the Qwen3-VL~\cite{bai2025qwen3} family spanning 8B to 235B parameters across Instruct and Think variants, Gemma-3~\cite{gemmateam2025gemma3technicalreport}, and MedGemma~\cite{sellergren2025medgemma}) deployed by Sglang~\cite{zheng2024sglang}. 
To explicitly decouple a model's intrinsic multimodal reasoning from its retrieval capability, we evaluate performance under four distinct evidence paradigms:
\begin{itemize}
    \item \textbf{Base:} The model receives only the raw query $Q$ including clinical history and corresponding images, measuring intrinsic diagnostic capability.
    \item \textbf{Physician Reference (PR):} We provide the abstracts of physician-cited references as oracle external context, establishing an upper bound.
    \item \textbf{RAG:} We ensemble bge-large-en-v1.5~\cite{xiao2024c} and MedCPT~\cite{jin2023medcpt} by retrieving top-5 candidates from each model and merging them with de-duplication (PMID-level), followed by score normalization and re-ranking.
    \item \textbf{Agent Retrieval (AR):} We adopt MiroFlow~\cite{team2025mirothinker} as an agentic retrieval pipeline that performs unconstrained, real-time web search.
\end{itemize}

\subsection{The Impact of Evidence Retrieval Paradigms}
\label{subsec:exp1_retrieval}

\begin{table}[tb]
\centering
\resizebox{\textwidth}{!}{%
\begin{tabular}{l *{12}{r}}
\toprule
\multirow{2}{*}{Model} & \multicolumn{3}{c}{History \& Images} & \multicolumn{3}{c}{w/ Physician Reference} & \multicolumn{3}{c}{w/ RAG} & \multicolumn{3}{c}{w/ Agent Retrieval} \\
\cmidrule(lr){2-4} \cmidrule(lr){5-7} \cmidrule(lr){8-10} \cmidrule(lr){11-13}
& Hit@1 & Hit@3 & MCQ & Hit@1 & Hit@3 & MCQ & Hit@1 & Hit@3 & MCQ & Hit@1 & Hit@3 & MCQ \\
\midrule
\multicolumn{13}{l}{\textit{Closed-source}} \\
GPT-5.2 & 25.4 & 35.0 & 87.4 & 73.4 & 81.8 & 86.8 & 15.4 & 23.6 & 85.6 & 25.5 & 41.1 & 87.4 \\
Gemini-3-Pro-Preview & 34.2 & 47.4 & 88.4 & 70.6 & 80.0 & 88.0 & 28.4 & 41.0 & 88.0 & 36.9 & 50.5 & 87.6 \\
\midrule
\multicolumn{13}{l}{\textit{Open-source}} \\
Kimi-K2.5 & 23.2 & 33.6 & 87.2 & 68.8 & 77.4 & 85.0 & 16.8 & 25.2 & 83.4 & 31.9 & 43.5 & 85.8 \\
GLM-4.6V & 10.8 & 21.4 & 78.2 & 61.6 & 77.0 & 78.8 & 7.2 & 13.0 & 77.4 & 16.8 & 31.7 & 76.2 \\
Qwen3-VL-235B-A22B-Inst & 13.0 & 21.6 & 78.6 & 69.8 & 79.4 & 79.8 & 10.2 & 17.6 & 77.6 & 16.7 & 29.4 & 78.6 \\
Qwen3-VL-235B-A22B-Think & 13.2 & 24.8 & 80.4 & 67.6 & 79.4 & 80.6 & 10.2 & 19.4 & 79.8 & 19.6 & 32.3 & 79.0 \\
Qwen3-VL-32B-Inst & 11.2 & 19.6 & 74.8 & 70.2 & 80.4 & 74.0 & 9.2 & 17.6 & 76.2 & 17.4 & 31.3 & 74.7 \\
Qwen3-VL-32B-Think & 12.6 & 20.2 & 78.2 & 65.0 & 75.8 & 78.4 & 11.8 & 19.2 & 73.8 & 17.8 & 31.3 & 75.6 \\
Qwen3-VL-30B-A3B-Inst & 7.6 & 14.8 & 75.8 & 66.4 & 76.0 & 76.0 & 8.4 & 14.4 & 75.6 & 18.0 & 31.1 & 74.9 \\
Qwen3-VL-30B-A3B-Think & 10.4 & 20.0 & 76.0 & 58.0 & 72.0 & 77.0 & 9.0 & 16.2 & 73.0 & 17.0 & 29.9 & 74.3 \\
Qwen3-VL-8B-Inst & 5.0 & 9.8 & 72.0 & 67.6 & 76.2 & 71.6 & 6.8 & 12.2 & 69.2 & 15.8 & 30.9 & 68.3 \\
Qwen3-VL-8B-Think & 7.8 & 14.0 & 70.6 & 56.0 & 72.2 & 70.8 & 7.2 & 13.0 & 67.6 & 17.2 & 30.7 & 68.5 \\
Gemma-3-12b-it & 8.8 & 20.8 & 65.2 & 79.2 & 90.0 & 65.2 & 10.2 & 18.4 & 63.6 & 16.4 & 29.7 & 65.1 \\
Gemma-3-4b-it & 6.8 & 14.2 & 53.6 & 64.4 & 81.2 & 54.6 & 7.4 & 15.4 & 47.8 & 16.4 & 29.7 & 48.5 \\
\midrule
\multicolumn{13}{l}{\textit{Medical Specific}} \\
MedGemma-27b-it & 15.8 & 26.8 & 75.2 & 64.2 & 85.0 & 74.0 & 14.6 & 28.6 & 72.4 & 17.6 & 31.3 & 69.9 \\
MedGemma-4b-it & 9.8 & 17.6 & 63.4 & 61.0 & 74.6 & 62.0 & 9.4 & 16.8 & 58.2 & 16.0 & 29.9 & 61.1 \\
\bottomrule
\end{tabular}%
}
\caption{Performance comparison of various models on the test set ($n=500$). The Base setting includes only clinical history and images as input, whereas the other configurations further augment this input with external context.}
\label{tab:performance_retrieval}
\end{table}

Table~\ref{tab:performance_retrieval} indicates that the effect of external evidence on multimodal clinical diagnostic reasoning is highly contingent on evidence quality, rather than being uniformly beneficial. When physician reference context is provided, all models exhibit substantial gains, suggesting that current systems already possess strong potential for medical logical deduction. In this sense, a major bottleneck in real-world deployment may not be the models’ reasoning capability per se, but the limited accessibility of high-quality, low-noise, diagnostic-level pathological associations. This trend holds for both proprietary and lightweight open-source models, highlighting the generality of the observation.

In contrast, under practical settings without human curation, the standard RAG paradigm fails to deliver the expected improvement and instead leads to an overall degradation. General-purpose retrieval tends to introduce redundant or weakly relevant literature, which effectively becomes “retrieval noise” that disrupts multimodal attention allocation and interferes with the models’ reasoning chains. Notably, MCQ accuracy exhibits relatively minor changes under RAG, whereas open-ended differential diagnosis (Hit@1/Hit@3) degrades more consistently, implying that noise primarily harms tasks requiring deep cross-evidence integration. This relative stability likely reflects an option-priming ceiling effect, where answer choices already activate internal knowledge, making MCQ less sensitive to retrieval noise. These results underscore that retrieval utility depends less on the quantity of retrieved text and more on whether it contains decisive, diagnosis-specific cues.

Compared with standard RAG, the Agent Retrieval paradigm shows promise in noise filtering and knowledge refinement via multi-step planning and iterative searching, recovering performance for most models and in several cases surpassing the baseline. However, a clear gap remains between Agent Retrieval and the upper bound achieved with physician reference context. This suggests that, while agentic retrieval can reduce irrelevant interference, current systems still struggle to reliably pinpoint the most discriminative diagnostic evidence from vast biomedical literature. The challenge is particularly pronounced for highly specialized, long-tail radiological or pathological features, where precise cue matching and evidence prioritization remain difficult.

\subsection{Analysis of Retrieved Evidence Quality}

\begin{figure}[tb]
    \centering
    \begin{minipage}{0.50\textwidth}
        \centering
        \includegraphics[width=\textwidth]{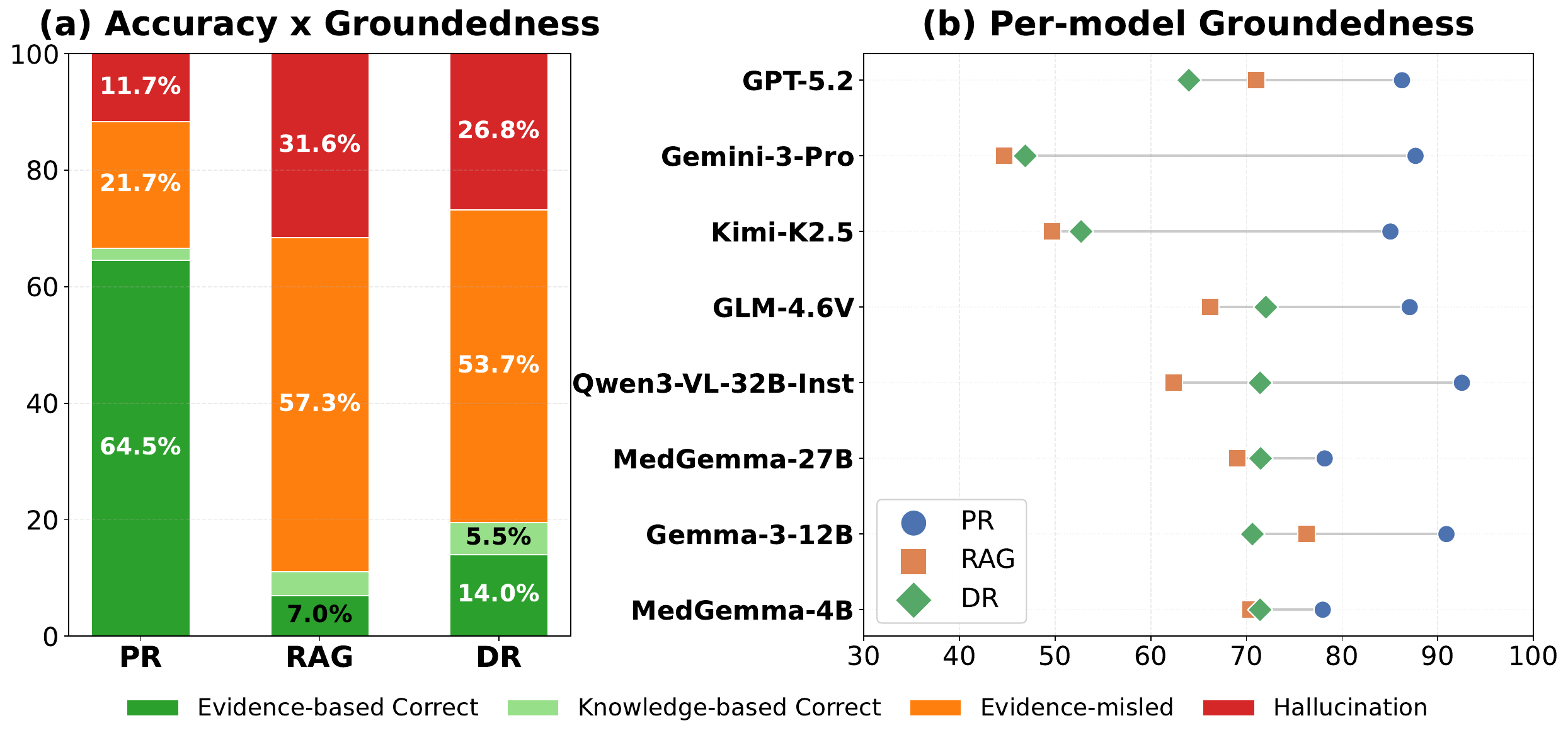}
        \caption{Faithfulness analysis across evidence conditions. (a) Distribution of accuracy × groundedness quadrants averaged over 16 models. (b) 
  Per-model groundedness rate under each evidence condition.}
        \label{fig:simi_b}
    \end{minipage}
    \hfill
    \begin{minipage}{0.45\textwidth}
        \centering
        \includegraphics[width=\textwidth]{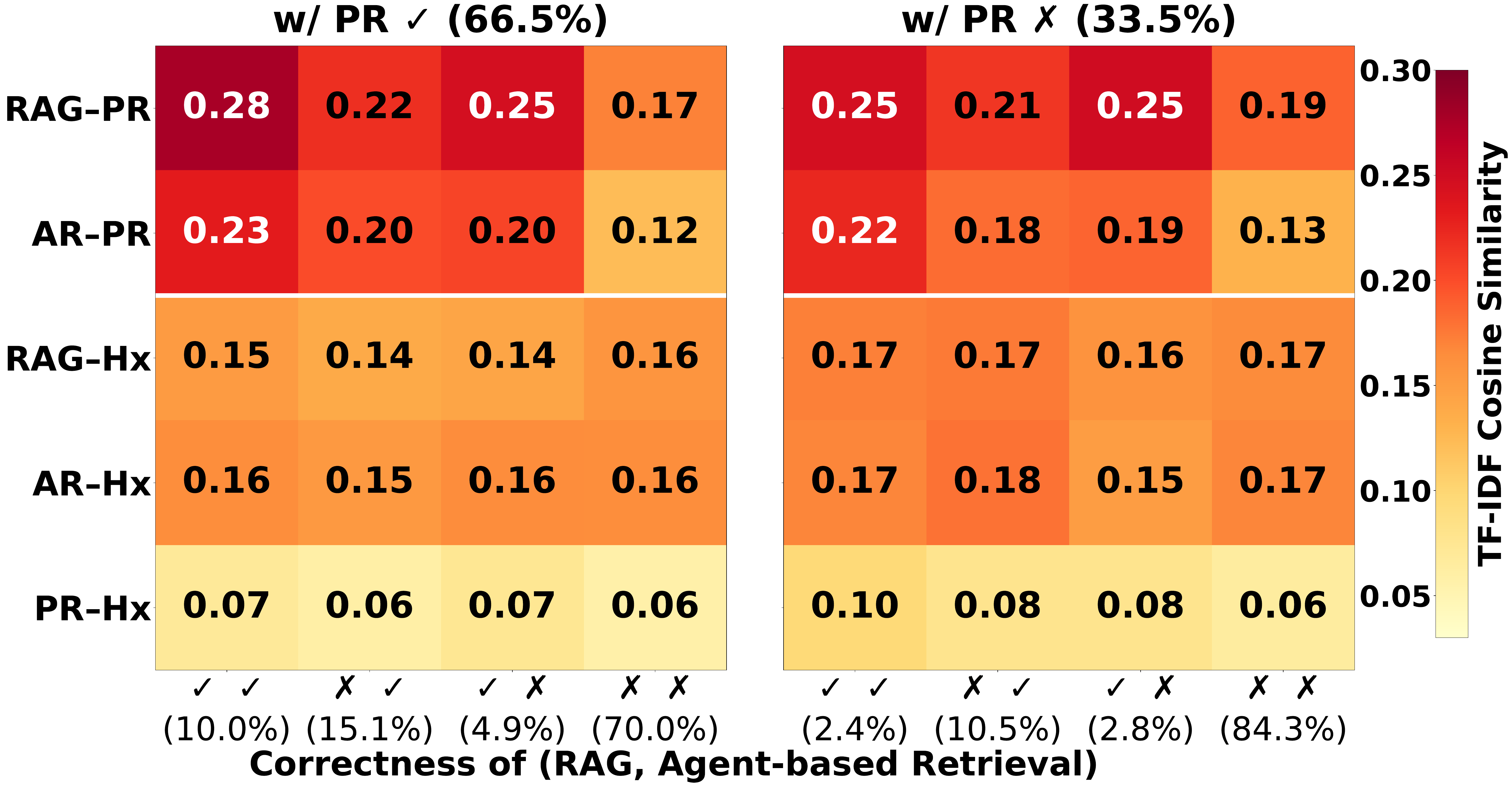}
        \caption{Heatmap of TF-IDF cosine similarities comparing retrieved evidence (RAG and AR), clinical history (Hx), and physician-annotated reference evidence (PR). Panels are stratified by baseline correctness.}
        \label{fig:simi_a}
    \end{minipage}
\end{figure}

Figure~\ref{fig:simi_b} shows that diagnostic behavior shifts dramatically with evidence quality. Under physician reference (coverage 92.1\%), most diagnoses are \emph{evidence-based correct} (64.5\%), indicating effective use of high-quality evidence. Under RAG (coverage 23.0\%), the dominant outcome becomes \emph{evidence-misled} (57.3\%) with high hallucination (31.6\%); Agent Retrieval improves coverage to 38.2\% and increases evidence-based correctness (7.0\%$\rightarrow$14.0\%), but \emph{evidence-misled} remains prevalent (53.7\%) and hallucination stays substantial (26.8\%). Meanwhile, \emph{knowledge-based correct} is consistently rare (2.1\%/4.1\%/5.5\% for PR/RAG/AR), suggesting models seldom override provided evidence using internal knowledge.

The core bottleneck in RAG/AR pipelines is retrieval quality and coverage rather than model unfaithfulness: models are often \emph{too} faithful to the retrieved documents, so when retrieval fails to include diagnostically relevant literature, grounded but incorrect reasoning becomes the dominant failure mode. Per-model analysis further indicates faithfulness is a double-edged sword—high groundedness can yield strong performance under curated evidence yet increases susceptibility to being misled under low-quality retrieval, whereas lower groundedness does not improve accuracy and instead correlates with higher hallucination.

Figure~\ref{fig:simi_a} provides a mechanistic explanation: diagnostic correctness positively tracks the lexical alignment between retrieved evidence and the physician-annotated evidence (e.g., within the PR $\checkmark$ cohort, RAG--PR similarity drops from 0.28 to 0.17 as outcomes worsen), while similarity to the clinical history remains largely uninformative (0.14--0.18 across strata). The intrinsically low Hx–PR similarity of 0.06 to 0.09 highlights a semantic gap between patient narratives and generalized medical knowledge, reinforcing that locating PR-like literature rather than matching Hx drives correct diagnosis.

\subsection{Impact of Input Modalities on Diagnostic Performance}

\begin{table}[tb]
\centering
\resizebox{\textwidth}{!}{%
\begin{tabular}{l *{12}{r}}
\toprule
\multirow{2}{*}{Model} & \multicolumn{3}{c}{History} & \multicolumn{3}{c}{w/ Image} & \multicolumn{3}{c}{w/ Findings} & \multicolumn{3}{c}{w/ Both} \\
\cmidrule(lr){2-4} \cmidrule(lr){5-7} \cmidrule(lr){8-10} \cmidrule(lr){11-13}
& Hit@1 & Hit@3 & MCQ & Hit@1 & Hit@3 & MCQ & Hit@1 & Hit@3 & MCQ & Hit@1 & Hit@3 & MCQ \\
\midrule
\multicolumn{13}{l}{\textit{Closed-source}} \\
GPT-5.2 & 16.4 & 26.2 & 87.0 & 25.4 & 35.0 & 87.4 & 54.2 & 64.6 & 90.0 & 55.6 & 65.8 & 91.0 \\
Gemini-3-Pro-Preview & 19.0 & 31.0 & 89.8 & 34.2 & 47.4 & 88.4 & 51.6 & 58.6 & 94.2 & 57.8 & 65.0 & 91.8 \\
\midrule
\multicolumn{13}{l}{\textit{Open-source}} \\
Kimi-K2.5 & 11.6 & 19.8 & 85.0 & 23.2 & 33.6 & 87.2 & 51.6 & 64.2 & 90.2 & 53.8 & 63.2 & 90.8 \\
GLM-4.6V & 8.6 & 19.4 & 78.2 & 10.8 & 21.4 & 78.2 & 41.8 & 54.8 & 84.0 & 41.2 & 57.2 & 84.4 \\
Qwen3-VL-235B-A22B-Inst & 8.4 & 14.8 & 78.6 & 13.0 & 21.6 & 78.6 & 46.2 & 60.4 & 85.2 & 49.0 & 58.6 & 87.6 \\
Qwen3-VL-235B-A22B-Think & 9.0 & 18.6 & 79.6 & 13.2 & 24.8 & 80.4 & 43.0 & 56.0 & 86.4 & 42.4 & 56.0 & 85.6 \\
Qwen3-VL-32B-Inst & 7.6 & 14.8 & 78.2 & 11.2 & 19.6 & 74.8 & 43.0 & 54.6 & 82.6 & 41.6 & 53.0 & 83.0 \\
Qwen3-VL-32B-Think & 10.6 & 18.0 & 77.6 & 12.6 & 20.2 & 78.2 & 38.0 & 49.0 & 83.4 & 41.8 & 50.6 & 83.8 \\
Qwen3-VL-30B-A3B-Inst & 7.0 & 13.0 & 77.0 & 7.6 & 14.8 & 75.8 & 40.6 & 54.8 & 83.0 & 38.6 & 51.6 & 82.4 \\
Qwen3-VL-30B-A3B-Think & 6.6 & 13.4 & 76.8 & 10.4 & 20.0 & 76.0 & 36.0 & 48.6 & 84.2 & 30.4 & 45.0 & 83.2 \\
Qwen3-VL-8B-Inst & 4.4 & 11.6 & 72.0 & 5.0 & 9.8 & 72.0 & 39.8 & 52.0 & 80.0 & 39.2 & 49.8 & 81.0 \\
Qwen3-VL-8B-Think & 8.4 & 13.0 & 71.4 & 7.8 & 14.0 & 70.6 & 34.6 & 44.0 & 80.8 & 34.4 & 43.4 & 79.6 \\
Gemma-3-12B & 9.0 & 19.2 & 66.6 & 8.8 & 20.8 & 65.2 & 43.4 & 59.0 & 77.6 & 44.4 & 60.4 & 77.0 \\
Gemma-3-4B & 6.8 & 15.0 & 54.4 & 6.8 & 14.2 & 53.6 & 41.2 & 58.2 & 66.2 & 39.2 & 58.0 & 66.2 \\
\midrule
\multicolumn{13}{l}{\textit{Medical Specific}} \\
MedGemma-27B-it & 13.4 & 22.8 & 73.4 & 15.8 & 26.8 & 75.2 & 45.2 & 62.2 & 81.8 & 46.6 & 63.0 & 81.4 \\
MedGemma-4B-it & 9.8 & 16.0 & 61.4 & 9.8 & 17.6 & 63.4 & 35.8 & 51.0 & 72.0 & 36.4 & 50.6 & 73.8 \\
\bottomrule
\end{tabular}%
}
\caption{Effect of input modalities on diagnostic performance on the test set. The Base setting includes only clinical history as input, whereas the other configurations further augment this input with additional modality information.}
\label{tab:performance_modal}
\end{table}

Table~\ref{tab:performance_modal} show a consistent performance ladder across all models under the four input settings. Clinical history only yields the lowest performance, and additional evidence progressively improves diagnosis. Adding textual findings (w/ Findings) produces large and consistent gains, improving both open-ended Hit@k and MCQ, which suggests that models can effectively leverage structured textual evidence for differential diagnosis. In contrast, adding images alone (w/ Image) provides smaller and more variable benefits that are mainly reflected in modest Hit@k increases, while MCQ shows no consistent improvement and even degrades for several models, indicating that candidate retrieval improves more reliably than final discrimination. The full multimodal setting (w/ both) generally achieves the best results and exhibits clearer synergy in a subset of high-capacity models.

These results indicate that image information is not useless, yet current multimodal LMs do not consistently convert visual cues into decision-level gains. The cues extracted from images appear sufficient to modestly improve retrieval-style metrics but remain unreliable under stronger discriminative constraints, consistent with the observed MCQ variability and regressions. By comparison, textual findings provide more direct and compositional diagnostic evidence, aligning with LLMs’ strengths in language-based reasoning. The overall best performance under w/ both and the additional gains observed in a few models even when findings are present suggest a clear headroom. With more reliable lesion localization, attribute extraction, and tighter alignment between imaging patterns and clinical semantics, images could shift from weak add-ons to strong evidence, improving discrimination beyond findings and substantially raising the ceiling when findings are unavailable.

\section{Conclusion}
We introduce CURE, a multimodal benchmark of 500 physician-annotated clinical cases that separates end-to-end diagnosis into evidence retrieval and clinical understanding. Across 16 models and four evidence paradigms, we find three key results: (1) retrieval rather than reasoning is the main deployment bottleneck, as models struggle to surface decisive literature at scale; (2) greater faithfulness improves performance with high-quality evidence but increases vulnerability to noisy or irrelevant retrieval; (3) textual findings consistently enhance diagnostic accuracy, whereas images provide only modest and variable gains, indicating substantial room for improvement in multimodal clinical understanding.

%
%
%
\bibliographystyle{splncs04}
\bibliography{reference}

@article{sellergren2025medgemma,
  title={Medgemma technical report},
  author={Sellergren, Andrew and Kazemzadeh, Sahar and Jaroensri, Tiam and Kiraly, Atilla and Traverse, Madeleine and Kohlberger, Timo and Xu, Shawn and Jamil, Fayaz and Hughes, C{\'\i}an and Lau, Charles and others},
  journal={arXiv preprint arXiv:2507.05201},
  year={2025}
}

@article{li2023llava,
  title={Llava-med: Training a large language-and-vision assistant for biomedicine in one day},
  author={Li, Chunyuan and Wong, Cliff and Zhang, Sheng and Usuyama, Naoto and Liu, Haotian and Yang, Jianwei and Naumann, Tristan and Poon, Hoifung and Gao, Jianfeng},
  journal={Advances in Neural Information Processing Systems},
  volume={36},
  pages={28541--28564},
  year={2023}
}

@article{huang2025elicit,
  title={Elicit and enhance: Advancing multimodal reasoning in medical scenarios},
  author={Huang, Zhongzhen and Mu, Linjie and Zhu, Yakun and Zhao, Xiangyu and Zhang, Shaoting and Zhang, Xiaofan},
  journal={arXiv preprint arXiv:2505.23118},
  year={2025}
}

@article{qin2026incentivizing,
  title={Incentivizing Cardiologist-Like Reasoning in MLLMs for Interpretable Echocardiographic Diagnosis},
  author={Qin, Yi and Wang, Lehan and Zhao, Chenxu and Lee, Alex PW and Li, Xiaomeng},
  journal={arXiv preprint arXiv:2601.08440},
  year={2026}
}

@article{huang2020fusion,
  title={Fusion of medical imaging and electronic health records using deep learning: a systematic review and implementation guidelines},
  author={Huang, Shih-Cheng and Pareek, Anuj and Seyyedi, Saeed and Banerjee, Imon and Lungren, Matthew P},
  journal={NPJ digital medicine},
  volume={3},
  number={1},
  pages={136},
  year={2020},
  publisher={Nature Publishing Group UK London}
}

@inproceedings{zhang2025multimodal,
  title={A Multimodal Contrastive Learning for Detecting Aortic Dissection on 3D Non-contrast CT with Anatomy Simplification},
  author={Zhang, Duoer and Xiao, Wenbo and Jiang, Chen and Qiu, Yuxuan and Feng, Zhan and Wang, Hong and Zheng, Yefeng and Zhu, Wentao},
  booktitle={International Conference on Medical Image Computing and Computer-Assisted Intervention},
  pages={3--12},
  year={2025},
  organization={Springer}
}

@article{del2014clinical,
  title={Clinical questions raised by clinicians at the point of care: a systematic review},
  author={Del Fiol, Guilherme and Workman, T Elizabeth and Gorman, Paul N},
  journal={JAMA internal medicine},
  volume={174},
  number={5},
  pages={710--718},
  year={2014}
}

@article{lewis2020retrieval,
  title={Retrieval-augmented generation for knowledge-intensive nlp tasks},
  author={Lewis, Patrick and Perez, Ethan and Piktus, Aleksandra and Petroni, Fabio and Karpukhin, Vladimir and Goyal, Naman and K{\"u}ttler, Heinrich and Lewis, Mike and Yih, Wen-tau and Rockt{\"a}schel, Tim and others},
  journal={Advances in neural information processing systems},
  volume={33},
  pages={9459--9474},
  year={2020}
}

@inproceedings{xiong2024benchmarking,
  title={Benchmarking retrieval-augmented generation for medicine},
  author={Xiong, Guangzhi and Jin, Qiao and Lu, Zhiyong and Zhang, Aidong},
  booktitle={Findings of the Association for Computational Linguistics: ACL 2024},
  pages={6233--6251},
  year={2024}
}

@article{team2025mirothinker,
  title={Mirothinker: Pushing the performance boundaries of open-source research agents via model, context, and interactive scaling},
  author={Team, MiroMind and Bai, Song and Bing, Lidong and Chen, Carson and Chen, Guanzheng and Chen, Yuntao and Chen, Zhe and Chen, Ziyi and Dai, Jifeng and Dong, Xuan and others},
  journal={arXiv preprint arXiv:2511.11793},
  year={2025}
}

@article{xu2025comprehensive,
  title={A comprehensive survey of deep research: Systems, methodologies, and applications},
  author={Xu, Renjun and Peng, Jingwen},
  journal={arXiv preprint arXiv:2506.12594},
  year={2025}
}

@inproceedings{zuo2025medxpertqa,
  title={MedXpertQA: Benchmarking Expert-Level Medical Reasoning and Understanding},
  author={Zuo, Yuxin and Qu, Shang and Li, Yifei and Chen, Zhang-Ren and Zhu, Xuekai and Hua, Ermo and Zhang, Kaiyan and Ding, Ning and Zhou, Bowen},
  booktitle={International Conference on Machine Learning},
  pages={80961--80990},
  year={2025},
  organization={PMLR}
}

@inproceedings{yue2024mmmu,
  title={Mmmu: A massive multi-discipline multimodal understanding and reasoning benchmark for expert agi},
  author={Yue, Xiang and Ni, Yuansheng and Zhang, Kai and Zheng, Tianyu and Liu, Ruoqi and Zhang, Ge and Stevens, Samuel and Jiang, Dongfu and Ren, Weiming and Sun, Yuxuan and others},
  booktitle={Proceedings of the IEEE/CVF conference on computer vision and pattern recognition},
  pages={9556--9567},
  year={2024},
}

@inproceedings{xiao2024c,
  title={C-pack: Packed resources for general chinese embeddings},
  author={Xiao, Shitao and Liu, Zheng and Zhang, Peitian and Muennighoff, Niklas and Lian, Defu and Nie, Jian-Yun},
  booktitle={Proceedings of the 47th international ACM SIGIR conference on research and development in information retrieval},
  pages={641--649},
  year={2024},
}

@article{jin2023medcpt,
  title={Medcpt: Contrastive pre-trained transformers with large-scale pubmed search logs for zero-shot biomedical information retrieval},
  author={Jin, Qiao and Kim, Won and Chen, Qingyu and Comeau, Donald C and Yeganova, Lana and Wilbur, W John and Lu, Zhiyong},
  journal={Bioinformatics},
  volume={39},
  number={11},
  pages={btad651},
  year={2023},
  publisher={Oxford University Press}
}

@article{singh2025openai,
  title={Openai gpt-5 system card},
  author={Singh, Aaditya and Fry, Adam and Perelman, Adam and Tart, Adam and Ganesh, Adi and El-Kishky, Ahmed and McLaughlin, Aidan and Low, Aiden and Ostrow, AJ and Ananthram, Akhila and others},
  journal={arXiv preprint arXiv:2601.03267},
  year={2025}
}

@misc{gemmateam2025gemma3technicalreport,
      title={Gemma 3 Technical Report}, 
      author={Gemma Team},
      year={2025},
      eprint={2503.19786},
      archivePrefix={arXiv},
      primaryClass={cs.CL},
      url={https://arxiv.org/abs/2503.19786}, 
}

@article{bai2025qwen3,
  title={Qwen3-vl technical report},
  author={Bai, Shuai and Cai, Yuxuan and Chen, Ruizhe and Chen, Keqin and Chen, Xionghui and Cheng, Zesen and Deng, Lianghao and Ding, Wei and Gao, Chang and Ge, Chunjiang and others},
  journal={arXiv preprint arXiv:2511.21631},
  year={2025}
}

@article{team2026kimi,
  title={Kimi K2. 5: Visual Agentic Intelligence},
  author={Team, Kimi and Bai, Tongtong and Bai, Yifan and Bao, Yiping and Cai, SH and Cao, Yuan and Charles, Y and Che, HS and Chen, Cheng and Chen, Guanduo and others},
  journal={arXiv preprint arXiv:2602.02276},
  year={2026}
}

@misc{vteam2025glm45vglm41vthinkingversatilemultimodal,
      title={GLM-4.5V and GLM-4.1V-Thinking: Towards Versatile Multimodal Reasoning with Scalable Reinforcement Learning},
      author={V Team},
      year={2025},
      eprint={2507.01006},
      archivePrefix={arXiv},
      primaryClass={cs.CV},
      url={https://arxiv.org/abs/2507.01006},
}

@article{zheng2024sglang,
  title={Sglang: Efficient execution of structured language model programs},
  author={Zheng, Lianmin and Yin, Liangsheng and Xie, Zhiqiang and Sun, Chuyue Livia and Huang, Jeff and Yu, Cody Hao and Cao, Shiyi and Kozyrakis, Christos and Stoica, Ion and Gonzalez, Joseph E and others},
  journal={Advances in neural information processing systems},
  volume={37},
  pages={62557--62583},
  year={2024},
}

\end{document}